\documentclass{article} % For LaTeX2e
\usepackage{iclr2027_conference,times}

\usepackage{amsmath,amsfonts,bm}

\def\eqref#1{equation~\ref{#1}}
\def\1{\bm{1}}

\DeclareMathAlphabet{\mathsfit}{\encodingdefault}{\sfdefault}{m}{sl}
\SetMathAlphabet{\mathsfit}{bold}{\encodingdefault}{\sfdefault}{bx}{n}

\usepackage{graphicx}
\usepackage{wrapfig}
\usepackage[T1]{fontenc}
\usepackage{booktabs}

\newcommand{\tablebodyfont}{%
  \ttfamily
  \renewcommand{\arraystretch}{1.08}%
}

\usepackage{xcolor} % Required for inserting images

\definecolor{mylinks}{RGB}{41, 171, 226} % New Colour
\definecolor{bestred}{RGB}{178, 24, 43}  % best score
\usepackage[colorlinks=true,urlcolor=mylinks,linkcolor=mylinks,citecolor=mylinks
]{hyperref}
\usepackage{natbib}
\usepackage{url}

\title{Do Better Goal Representations Improve Goal-Conditioned Reinforcement Learning?}

\author{Syed Nazmus Sakib$^{1}$ \quad Abdul Monaf Chowdhury$^{1}$ \quad Nafiul Haque$^{1}$ \\
\textbf{Shifat E Arman$^{1,2}$ \quad Md Mehedi Hasan$^{1}$} \\
$^{1}$Department of Robotics and Mechatronics Engineering, University of Dhaka, Bangladesh \\
$^{2}$Department of Computer Science, University of Oxford, UK
}

\iclrfinalcopy % arXiv preprint: shows authors and removes line numbers
\begin{document}

\maketitle
\lhead{Preprint}

\begin{abstract}
Goal-conditioned reinforcement learning (GCRL) relies heavily on how target goals are represented to the policy. While recent methods encode goals via temporal distance, occupancy, or controllability, it remains unclear how much downstream performance actually depends on representation quality. We study this in offline GCRL by constructing an exact temporal-distance goal representation in deterministic mazes. We then systematically corrupt its geometric quality while keeping the downstream learner fixed. Across OGBench navigation tasks and two algorithms, large changes in goal-representation quality produce almost no change in performance. However, applying the same interventions to the agent's current state more than doubles success, revealing the state pathway as the true bottleneck. Building on this insight, we show that simple random Fourier positional encodings substantially improve performance on the hardest navigation tasks without map information or objective modifications. Overall, our findings suggest that in state-based offline navigation, improving how the agent's current state is represented matters far more than refining the goal representation. Code will be released soon.
\end{abstract}

\section{Introduction}
\label{sec:introduction}

Goal-conditioned reinforcement learning (GCRL) learns a single policy for reaching diverse target states by
conditioning behaviour on a specified goal \citep{kaelbling1993ijcai-learning, 3045118.3045258}. Offline GCRL learns such
policies from a fixed dataset, often using hindsight relabelling to construct goal-conditioned training
examples \citep{3295222.3295258, ICLR2025_ecd92623}. A central question in this setting is how the goal
should be represented for downstream control. Prior work has therefore developed representations intended to
capture task-relevant relational structure beyond the raw goal observation.

Recent approaches characterise
goals through future occupancy, temporal distance, or controllability, including contrastive representations
\citep{3600270.3602850}, value-implicit pre-training \citep{ma2023vipuniversalvisualreward}, quasimetric and temporal-distance
embeddings \citep{wang2023optimalgoalreachingreinforcementlearning, park2024foundationpolicieshilbertrepresentations, NEURIPS2025_dc8fe792}, and dual goal representations \citep{ICLR2026_f8cd7eb1}. Despite differences in their objectives, these methods are motivated by a common premise: a goal
representation that more accurately captures reachability structure, particularly the temporal relationships
between states, should enable more effective goal-conditioned control.

However, existing evaluations do not isolate the effect of goal-representation quality on downstream control.
Prior comparisons typically vary the representation-learning procedure and the resulting embedding at the same time. This
makes it difficult to determine whether performance differences arise from the information encoded by the
representation or from other aspects of its learning process. We therefore ask a more direct question: 
% if the downstream agent were given an exact representation of goal reachability while the rest of the learning
% pipeline remained unchanged, how much would control performance improve?

\textbf{If a goal-conditioned agent were handed a perfect goal representation, how much better would it act?}

Dual Goal Representations (DGR) provides such a setting by characterising goals through temporal distances and
establishing their sufficiency for optimal control \citep{ICLR2026_f8cd7eb1}. In deterministic maze environments,
these temporal distances correspond to shortest-path distances in the underlying transition graph and can
therefore be computed directly. This allows us to construct the DGR representation from exact distances, which we term the ideal representation, and
replace its learned approximation while leaving the downstream algorithm, training objective, and all other
components unchanged. 

\begin{figure}[!t]
\centering
\includegraphics[width=\textwidth]{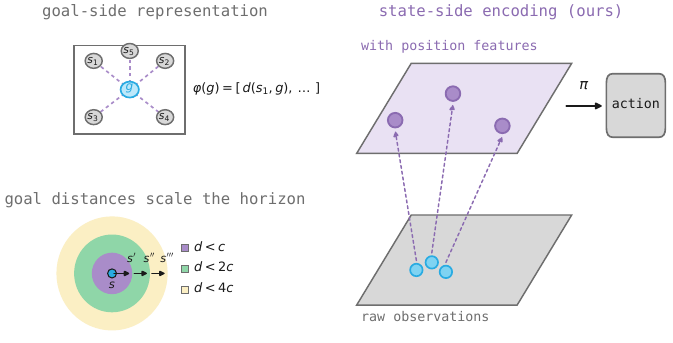}
\caption{\textbf{Two sides of the same interface.} \emph{Left, top:} goal-side representations, the axis this
literature optimises. A goal is described by its temporal distances to other states. \emph{Left, bottom:} the property that motivates them, namely that such
distances compose, so that a policy accurate within one radius extends to two and four
\citep{ICLR2025_bddd4e76}.
\emph{Right:} our proposal. We encode the agent's own position instead, lifting
observations that are close together in raw coordinates into a representation in which they are far apart, and
from which the policy acts.}
\label{fig:concept}
\end{figure}

We then systematically vary the quality of this representation to determine whether
downstream control depends on the reachability information it encodes. The resulting analysis shows that learned goal-side representations achieve success rates comparable to the exact ideal representation. We therefore hypothesise that the state pathway, through which the agent's current state is presented to the policy, is the more critical bottleneck. The remainder of the paper focuses on developing a matched set of state-side interventions to test this hypothesis. Figure~\ref{fig:concept} contrasts the two sides of the interface.
\begin{wrapfigure}{r}{0.5\textwidth}
    \centering
    \vspace{-10pt}
    \includegraphics[scale=0.65]{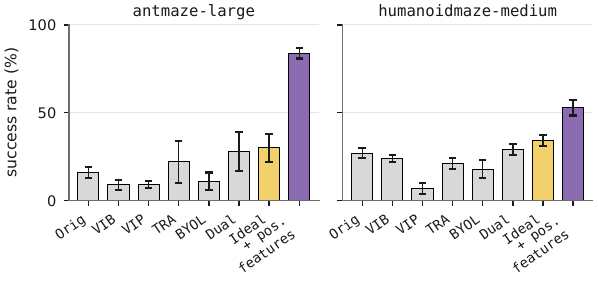}
\caption{\textbf{Goal representations are saturated; the state pathway is not.} GCIVL on antmaze-large. \emph{Grey:} published goal methods. \emph{Yellow:} exact BFS goal representation. \emph{Violet:} adding unprivileged state position features.}
\label{fig:teaser}
\vspace{-10pt}
\end{wrapfigure}
We propose a minimal method where random Fourier features of the agent's coordinates are appended to the observation. The method requires no maze layout and no modification to the learning objective. It only requires identifying which observation dimensions correspond to the agent's position. Despite its simplicity, it raises GCIVL success on antmaze-large from 32 to 84 and it roughly doubles success on humanoidmaze-medium (Figure~\ref{fig:teaser}). We show that with the downstream algorithm held fixed, the input pathway matters considerably more than the goal representation.

Our main \textbf{contributions} are threefold. First, we give the first measurement of an exactly computed ideal goal representation on continuous control, together with a probe-free way to measure representation quality, and show that representation quality is not the binding constraint on these tasks. Second, we identify the state pathway, rather than the goal representation, as the bottleneck, and support this with a matched pair of experiments in which the same intervention does nothing on the goal side and doubles performance on the state side. Third, we propose position features, a simple method that requires no privileged information and substantially improves performance on the hardest navigation tasks we study.

\section{Preliminaries}
\label{sec:preliminaries}

\textbf{Offline goal-conditioned reinforcement learning.}
We consider a controlled Markov process with state space
$\mathcal{S}$, action space $\mathcal{A}$, and transition kernel $p(s' \mid s, a)$, together with a goal space
$\mathcal{G} \subseteq \mathcal{S}$. A goal-conditioned policy $\pi(a \mid s, g)$ is trained to reach $g$ from
$s$. Following the benchmark convention \citep{ICLR2025_ecd92623}, the agent receives a reward of $0$ on reaching the
goal and $-1$ otherwise, discounted by $\gamma$, so that maximising return is equivalent to minimising the
number of steps taken. Learning is offline: the agent has a fixed dataset $\mathcal{D}$ of trajectories
collected by an unknown behaviour policy and may not interact with the environment during training. Goals are
supplied by hindsight relabelling of states in $\mathcal{D}$ \citep{3295222.3295258}.

\textbf{Temporal distances and dual goal representations.} A natural way to structure goal representations is through reachability, quantified by temporal distance. Formally, let $d(s, g)$ denote the minimal expected time (or shortest-path cost) required for an agent to transition from state $s$ to goal $g$ under the environment dynamics. Dual Goal Representations \citep[DGR;][]{ICLR2026_f8cd7eb1} leverage this metric directly, representing a goal not by its raw visual or coordinate features, but by the collection of temporal distances from all states to that goal. The idealised representation is the function
\[
\varphi^{\vee}(g) \;:=\; \bigl(s \mapsto d(s, g)\bigr),
\]
which is provably sufficient for optimal control and invariant to dynamics that do not affect reachability. Because $\varphi^{\vee}(g)$ is infinite-dimensional, practical implementations instantiate it by evaluating distances against a finite set of $K$ landmark states $a_1, \dots, a_K$ sampled from the dataset:
\begin{equation}
\varphi(g) \;=\; \bigl[\, d(a_1, g), \; \dots, \; d(a_K, g) \,\bigr] \in \mathbb{R}^{K}.
\label{eq:goal-representation}
\end{equation}
In standard DGR, the entries of~\eqref{eq:goal-representation} cannot be accessed directly in continuous domains and are instead approximated via a learned bilinear value parameterisation trained offline, which is then fed into downstream policy learners with a stop-gradient. Throughout this paper, we refer to this parametric estimate as the \textbf{learned} representation, and contrast it with the \textbf{ideal} representation obtained by evaluating~\eqref{eq:goal-representation} using exact ground-truth distances.

\textbf{The goal interface.}
Any goal-conditioned network combines two arguments before producing an output. We make
that combination explicit and call it the goal interface,
\[
I : \mathcal{S} \times \mathbb{R}^{K} \to \mathbb{R}^{m}, \qquad
\pi(a \mid s, g) \;=\; h\bigl(I(s, \varphi(g))\bigr),
\]
where $h$ is the action head, and likewise for value networks. Every goal-representation method we are aware
of, including all of those we compare against, fixes the interface to concatenation followed by a multilayer
perceptron,
\begin{equation}
I_{\mathrm{concat}}(s, \varphi) \;=\; \mathrm{MLP}\bigl([\, s \, ; \, \varphi \,]\bigr).
\label{eq:concat-interface}
\end{equation}
The choice in~\eqref{eq:concat-interface} is inherited rather than argued for. The closest existing comparisons
vary something else: \citet{ICLR2026_f8cd7eb1} compare aggregation functions inside the representation-learning
objective, and compare early against late fusion for pixel encoders. Neither varies how a state-based goal
representation is combined with the state inside the downstream policy. The representation $\varphi$, the
interface $I$, and the state pathway are separate design axes, and this paper measures all three.
Appendix~\ref{app:interfaces} defines the interfaces we compare and Appendix~\ref{app:interface-sweep} reports
the full sweep.

% \textbf{Environments and evaluation protocol.}
% We evaluate on the state-based navigation tasks of OGBench \citep{ogbench}: pointmaze-medium, pointmaze-large, antmaze-medium, antmaze-large and humanoidmaze-medium. These
% are the tasks in which temporal distance is a graph geodesic, which is what makes the ideal representation
% computable. Each run trains for one million gradient steps. Following the benchmark protocol, we evaluate on
% five held-out goals with fifty episodes each, and we report the mean success rate over the final three
% evaluations, at 800k, 900k and one million steps. We never report the best evaluation over training. Arms are
% compared with Welch's unequal-variance $t$-test and 95\% confidence intervals. Because several of our central
% claims are null results, we state sample sizes in every table and report confidence intervals rather than
% $p$-values alone, following recommended practice for reinforcement learning evaluation \citep{deep-rl-that-matters, statistical-precipice}.

\textbf{Downstream algorithms.}
We use the two state-based algorithms reported by \citet{ICLR2026_f8cd7eb1}. GCIVL learns a
goal-conditioned value function by expectile regression and extracts a policy by advantage-weighted regression
\citep{kostrikov2021offlinereinforcementlearningimplicit}. CRL learns a contrastive critic that is bilinear in the state and goal embeddings and
extracts a policy with a behaviour-regularised deterministic objective \citep{3600270.3602850}. The two differ
in how they estimate value and in how they extract a policy, which lets us separate effects that are specific
to one family from those that are not. Appendix~\ref{app:architecture} gives both objectives, the networks and
every hyperparameter.

\section{Representation Interventions}
\label{sec:experimental-design}
This section outlines our framework for isolating the policy's input interface. We first construct an exact goal-reachability representation with a controlled corruption ladder, and then design a matched set of state-side interventions to test whether the observation pathway is the true bottleneck.

\subsection{Goal-Side Intervention}
\label{sec:constructing-ideal}

To isolate the effect of goal-representation quality, we construct an exact temporal-distance representation
in the deterministic navigation environments. For each landmark $a_i$, we compute the shortest-path distance
to every reachable maze cell using breadth-first search and represent a goal $g$ as
\[
\varphi(g) = \bigl[\, d(a_1,g), \ldots, d(a_K,g) \,\bigr] \in \mathbb{R}^K.
\]

We use the same landmark dimensionality as DGR, with $K=256$ for AntMaze and HumanoidMaze and $K=64$ for
PointMaze. The resulting representation is computed directly from the environment geometry rather than
learned from the offline dataset, while the downstream learning algorithm and all other training components
remain unchanged. Appendix~\ref{app:setup} describes the table construction and the evaluation protocol.

\paragraph{Measuring representation quality.}
\label{sec:representation-quality}
We quantify how well a representation preserves temporal-distance structure using min-plus decoding. Given a
state $s$ and representation $\varphi(g)$, we reconstruct the state-goal distance as
\begin{equation}
\hat d(s,g) = \min_i \left[ d(a_i,s) + \varphi_i(g) \right].
\label{eq:min-plus-decoding}
\end{equation}
Representation quality is measured by the Spearman correlation between $\hat d(s,g)$ and the true
shortest-path distance $d(s,g)$ over a fixed set of state-goal pairs. We additionally compute this correlation
within distance quartiles to distinguish local from long-range structure. This provides a direct measure of
the geometric information retained by the representation without fitting an auxiliary probe.
Appendix~\ref{app:quality} reports the correctness check on the decode and explains why a learned probe is
uninformative here.

\paragraph{Varying representation quality.}
\label{sec:representation-corruption}
Starting from the exact representation, we progressively add noise to its distance structure using two
perturbations. Gaussian perturbation modifies distances across the full range, whereas far-field perturbation
modifies only distances in the furthest quartile while preserving nearby relationships:
\begin{equation}
\tilde{\varphi}_i(g) = \max\left(0, \varphi_i(g) + \sigma\epsilon_{i,g}\right), \qquad
\epsilon_{i,g} \sim \mathcal{N}(0,1).
\label{eq:representation-corruption}
\end{equation}
The perturbation is sampled once and fixed throughout training, so each setting defines a deterministic
representation rather than stochastic observation noise. Varying $\sigma$ therefore produces representations
with progressively different amounts and types of temporal-distance information while leaving the downstream
learner unchanged.

\subsection{State-Side Intervention}
\label{sec:state-codes}

The goal-side experiments vary what information is provided about the target. We construct a matched
intervention on the agent state to determine whether the same representational structure has a different
effect when applied to the current observation. For each maze cell, we define a fixed state code
$c(s)\in\mathbb{R}^K$ and augment the observation as
\begin{equation}
\tilde{s} = \bigl[\, s; c(s) \,\bigr].
\label{eq:state-code}
\end{equation}
The code uses the same dimensionality and indexing as the goal representation. We evaluate exact
temporal-distance codes, corrupted variants, and a random code sampled independently for each cell. The random
code preserves state identity while containing no temporal-distance structure, allowing us to distinguish the
value of geometric structure from simply making the agent's position more distinguishable.

The table-based state codes require access to the maze geometry and are therefore used only as diagnostic
interventions. To obtain a representation that does not depend on the map, we instead encode the agent
coordinates $x,y$ using fixed random Fourier features. With $B\in\mathbb{R}^{F\times2}$,
\begin{equation}
\tilde{s} = \left[ s; \sin(2\pi B\,xy); \cos(2\pi B\,xy) \right].
\label{eq:position-features}
\end{equation}
We use $F=128$, adding 256 fixed features to the observation. The feature matrix is sampled once and remains
fixed across training, while the objective, optimiser, network widths, goal representation, and remaining
hyperparameters are unchanged. This provides a simple state encoding that can be applied without access to the
maze layout or transition graph.
\section{Goal Representations Are Saturated}
\label{sec:saturated}

This section examines how strongly downstream performance depends on goal-representation quality. We first replace the learned representation with an exact temporal-distance representation, then progressively reduce the distance information it preserves, and finally test a representation with no meaningful distance structure. Appendix~\ref{app:goal-side} reports the same comparisons on the remaining environments.

\begin{table}[!t]
\centering
\caption{\textbf{An exactly optimal goal representation does not reliably improve control.} GCIVL and CRL,
concatenation interface, success rate $\times 100$, mean $\pm$ standard deviation over $n$ seeds given in
parentheses. The published column is taken from \citet{ICLR2026_f8cd7eb1}, their Table 1 for GCIVL and Table 6
for CRL, over 8 seeds. The difference is ideal minus learned with a 95\% Welch interval; bold marks intervals excluding zero.}

\label{tab:ideal-vs-learned}
\footnotesize
\tablebodyfont
\setlength{\tabcolsep}{2.5pt} % Reduced column padding
\resizebox{\columnwidth}{!}{%   % Scales the table to fit the exact column width
\begin{tabular}{llcccc}
\toprule
Environment & Algo & Dual (published) & Learned (ours) & Ideal (ours) & Difference \\
\midrule
pointmaze-med & GCIVL & $76 \pm 7$ & $70.9 \pm 6.1$ (5) & $62.7 \pm 3.0$ (5) & $\boldsymbol{-8.2\;[-15.6, -0.7]}$ \\
pointmaze-large & GCIVL & $46 \pm 6$ & $46.4 \pm 6.5$ (5) & $46.3 \pm 7.9$ (5) & $-0.1\;[-10.7, +10.5]$ \\
antmaze-med & GCIVL & $75 \pm 4$ & $76.5 \pm 7.4$ (5) & $68.4 \pm 4.0$ (5) & $-8.0\;[-17.2, +1.1]$ \\
antmaze-large & GCIVL & $28 \pm 11$ & $32.2 \pm 8.9$ (8) & $30.0 \pm 8.0$ (8) & $-2.2\;[-11.3, +6.9]$ \\
humanoid-med & GCIVL & $29 \pm 3$ & $27.8 \pm 3.6$ (5) & $34.2 \pm 3.1$ (5) & $\boldsymbol{+6.4\;[+1.5, +11.4]}$ \\
\midrule
pointmaze-med & CRL & $33 \pm 1$ & $37.8 \pm 3.3$ (5) & $57.4 \pm 11.6$ (5) & $\boldsymbol{+19.5\;[+5.4, +33.7]}$ \\
pointmaze-large & CRL & $39 \pm 12$ & $35.4 \pm 7.9$ (5) & $43.0 \pm 17.6$ (4) & $+7.6\;[-18.8, +34.1]$ \\
antmaze-med & CRL & $93 \pm 3$ & $94.6 \pm 1.6$ (2) & $92.5 \pm 0.4$ (2) & $-2.1\;[-13.8, +9.7]$ \\
antmaze-large & CRL & $87 \pm 2$ & $82.2 \pm 3.4$ (8) & $80.8 \pm 3.4$ (8) & $-1.4\;[-5.1, +2.2]$ \\
\bottomrule
\end{tabular}%
}
\end{table}
\subsection{Does an Exact Goal Representation Improve Downstream Performance?}
\label{sec:ideal-control}

\begin{wrapfigure}{r}{0.5\textwidth}
    \centering
    \vspace{-10pt}
    \includegraphics[scale=0.65]{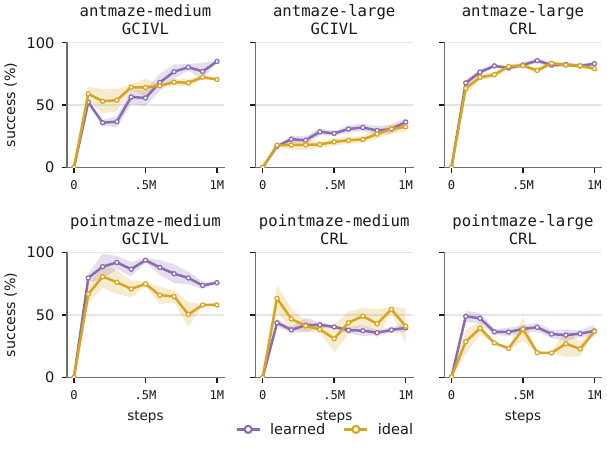}
    \caption{\textbf{Exact goal representations yield no consistent training advantage.} Evaluation curves (mean $\pm$ 1 SE over 2--8 seeds per panel; violet is learned, gold exact). Exact and learned representations exhibit indistinguishable learning rates and asymptotic success.}
    \label{fig:curves}
    \vspace{-20pt}
\end{wrapfigure}

We present the comparison between the learned dual representation and the exact temporal-distance
representation in Table~\ref{tab:ideal-vs-learned}, while keeping the downstream algorithm and all other training components fixed.

Of the nine evaluations, six show no statistically distinguishable difference between the two representations.
Among the remaining three, the exact representation improves success by 6.4 points on
humanoidmaze-medium with GCIVL and by 19.5 points on pointmaze-medium with CRL, whereas
the learned representation performs 8.2 points better on pointmaze-medium with GCIVL.
The differences therefore do not follow a consistent direction across environments or algorithms.

The same pattern is also reflected in the learning curves in Figure~\ref{fig:curves}: replacing the learned
representation with exact temporal distances does not consistently improve learning speed, final performance,
or the overall training trajectory. Across the five GCIVL tasks, the exact representation achieves a mean
success rate of 48, compared with 51 for the learned representation.

These results extend the evaluation of exact dual goal representations beyond the tabular setting considered
by \citet{ICLR2026_f8cd7eb1}. In their Lights-Out experiment, the exact representation improves over its
learned approximation, but we do not observe the same advantage consistently in the continuous-control
navigation tasks studied here. More accurate temporal-distance information therefore does not, by itself,
translate reliably into better downstream performance. This motivates a direct examination of whether the
downstream learner makes meaningful use of the distance structure encoded by the goal representation.

\subsection{How Sensitive Is Downstream Performance to Goal Representation Quality?}
\label{sec:quality-control}

\begin{figure}[!t]
    \centering
    \includegraphics[width=\textwidth]{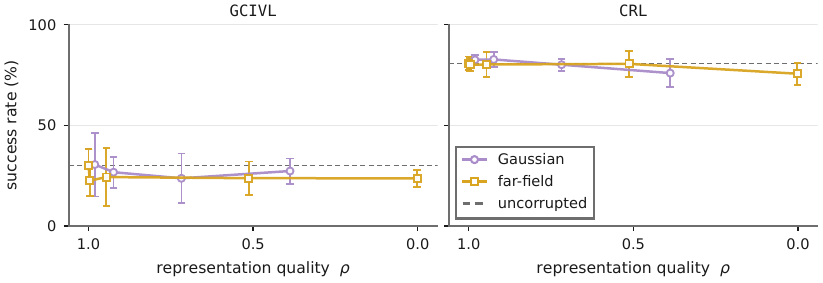}
   \caption{\textbf{Control is insensitive to goal quality.} Results on antmaze-large (mean $\pm$ 1 SD, 5--8 seeds). Across corruption ladders spanning exact geodesics to zero distance information, performance differences are statistically indistinguishable from zero.}
    \label{fig:ladder}
\end{figure}

A comparison between the learned and exact representations alone cannot determine whether representation quality matters, because the learned representation may already preserve enough of the relevant distance structure. We therefore evaluate the corruption ladder introduced in Section~\ref{sec:representation-corruption}, progressively reducing the quality of the exact representation while keeping the downstream learner unchanged. Representation quality is measured by the Spearman correlation between the min-plus decoded distance and the true geodesic distance.

Despite spanning a wide range of representation quality, the corresponding changes in performance remain small. Under far-field corruption at $\sigma=8$, the representation loses essentially all rank information about distant goals, yet performance decreases by only 6.4 points for GCIVL and 5.1 points for CRL. Across all sixteen settings in Figure~\ref{fig:ladder}, every point estimate remains within 7.6 points of the uncorrupted representation, and all confidence intervals include zero. Performance also does not vary monotonically with representation quality: under GCIVL, far-field $\sigma=1$ produces the lowest success despite retaining the second-highest measured quality, while Gaussian $\sigma=8$ outperforms Gaussian $\sigma=4$. Across the corruption families, the correlation between measured representation quality and success remains between $+0.21$ and $+0.27$, with no relationships distinguishable from zero.

The same pattern holds when we distinguish between local and long-range distance information. Far-field corruption keeps the near-field correlation at 0.997 across all noise levels while reducing the far-field correlation from 1.00 to $-0.41$. At the strongest corruption level, nearby goals remain ordered almost perfectly while distant goals are ranked in reverse, yet the resulting performance change remains limited. 

For CRL, far-field corruption changes success by only 0.2 points at $\sigma=4$ and 5.1 points at $\sigma=8$. These results indicate that neither local nor long-range distance structure produces a strong or consistent effect on downstream performance in this setting. We do not claim exact flatness: our stated tolerance was five points, and several confidence intervals extend beyond that range. The evidence instead supports a more limited conclusion: across a corruption ladder spanning nearly the full range of measured representation quality, no performance difference is statistically distinguishable from zero, whereas the state-side interventions examined in Section~\ref{sec:state-pathway} produce changes of roughly 30 to 50 points.

\subsection{Does the Goal Representation Need to Encode Distance at All?}
\label{sec:random-goal}

The ladder degrades distances but preserves the construction. The sharper question is whether the downstream
agent uses distance information at all. We replace the goal representation with a table of the same shape and
indexing whose entries are drawn i.i.d.\ from $\mathcal{N}(0,1)$. It identifies the goal cell and contains no
geometry whatsoever: its rank correlation with the true distances is 0.004.

\subsection{The Same Code Is Worth Forty Points More on the State Side}
\label{sec:matched-pair}
\begin{wrapfigure}{r}{0.5\textwidth}
    \centering
    \vspace{-10pt}
    \includegraphics[width=\linewidth]{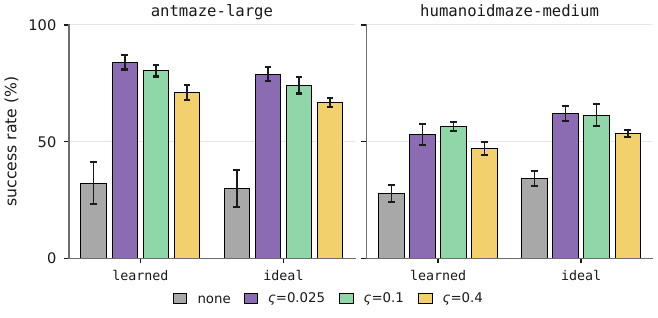}
\caption{\textbf{Position features.} Success with and without the features of
equation~\eqref{eq:position-features}, for both goal representations, at frequency scales
corresponding to wavelengths $8.5$, $2.1$ and $0.5$ maze cells.}
\label{fig:position-features}
\vspace{-35pt}
\end{wrapfigure}

That representation scores $31.9 \pm 2.4$ on antmaze-large, against $30.0 \pm 8.0$ for the exact geodesics,
$+1.9\;[-4.9, +8.8]$, and against $32.2 \pm 8.9$ for DGR's learned representation, $-0.2\;[-7.9, +7.4]$.
A random code, an exactly optimal code and a learned code are
indistinguishable (Figure~\ref{fig:state-and-goal}a). Whatever these agents extract from the goal, it is little more than its identity. This also
disposes of the ``already good enough'' reading of Section~\ref{sec:ideal-control}: a representation that is not
good at all does just as well.
% (Figure~\ref{fig:goal-vs-state}, upper rows).
\section{The State Pathway Is the Bottleneck}
\label{sec:state-pathway}

If the goal side is saturated, the headroom lies elsewhere. This section moves the identical intervention to
the other side of the interface, identifies what the state actually needs, and proposes a method
that requires no privileged information.
\begin{figure*}[!t]
\centering
\begin{minipage}[t]{0.48\textwidth}
\centering
\includegraphics[width=\linewidth]{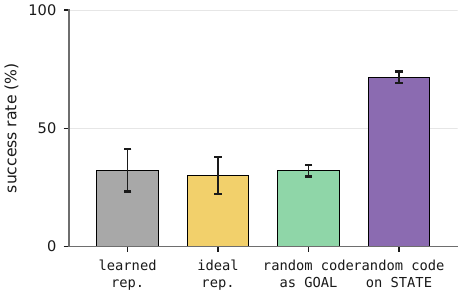}
\vspace{0.1cm}
\\(a)
\end{minipage}\hfill
\begin{minipage}[t]{0.48\textwidth}
\centering
\includegraphics[width=\linewidth]{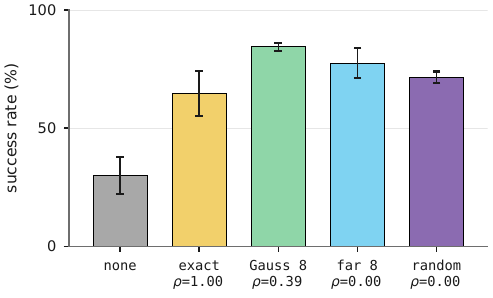}
\vspace{0.1cm}
\\(b)
\end{minipage}
\caption{\textbf{State-side interventions on antmaze-large (GCIVL; mean $\pm$ 1 SD, 5--8 seeds).} \textbf{(a)} Applying an identical fixed random $\mathcal{N}(0,1)$ code to the state more than doubles success, while doing nothing on the goal side. \textbf{(b)} The state code does not require distance geometry: corrupted and purely random codes match or outperform exact geodesics.}
\label{fig:state-and-goal}
\end{figure*}

Figure \ref{fig:state-and-goal}a presents a matched pair of interventions. The construction, the dimensionality, the cell indexing and the information content are identical in the last two rows of the figure; only the argument the code is applied to differs. The gap between these last two rows is 39.6 points, produced by nothing other than which argument receives the code. The same result holds for CRL, where an exact state code raises success from $80.8 \pm 3.4$ to $89.9 \pm 1.5$, a difference of $+9.2\;[+6.1, +12.3]$, on a task where the goal-side intervention was worth $-1.4\;[-5.1, +2.2]$. Appendix~\ref{app:state-side} tabulates this and the remaining state-side comparisons.

\subsection{What Must the State Code Contain?}
\label{sec:state-content}

Having found that the state side matters, we ask what it actually needs to contain. Figure~\ref{fig:state-and-goal}b explores this by running the state code through the same corruption ladder used for the goal, alongside the random table. Every code helps, and the exact geodesics are the weakest of them. Corrupting the table improves it, by 19.8 points for Gaussian $\sigma = 8$ and 12.9 points for far-field $\sigma = 8$, and a table with no geometry whatsoever is worth 6.8 points more than the exact one, $+6.8\;[-4.8, +18.3]$. The effect is therefore positional encoding: what the network gains is a high-dimensional, distinguishable representation of where it currently is, and the exact geodesic table is a comparatively poor one because it varies smoothly and nearly linearly across neighbouring cells.
\subsection{Position Features}
\label{sec:position-results}
While discrete state codes require access to the maze layout, the position features in \eqref{eq:position-features} require only the agent's raw coordinates. Figure~\ref{fig:position-features} evaluates this intervention on the two hardest navigation tasks across both goal representations and three frequency scales. 

On antmaze-large, the method raises GCIVL success from 32.2 to 83.9 when paired with DGR's learned representation. By comparison, published goal-representation methods span 9 to 28 on this task \citep[Table 1]{ICLR2026_f8cd7eb1}. On humanoidmaze-medium, it roughly doubles success under both representations. Across all settings, lower frequencies perform best, while the highest frequency performs worst (Figure~\ref{fig:frequency-trend}).

\begin{figure}[!t]
\centering
\includegraphics[width=\textwidth]{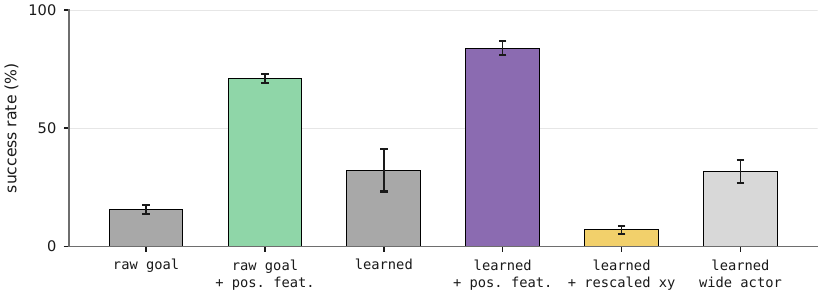}
\caption{\textbf{Controls for the method.} antmaze-large, GCIVL. Position features work without any goal
representation at all (second bar). Standardising the coordinates in place, which changes their scale without
adding a basis, is severely harmful (fifth bar), and widening the actor changes nothing (sixth bar).}
\label{fig:controls}
\vspace{-10pt}
\end{figure}

Ablations in Figure~\ref{fig:controls} isolate the source of this gain. Crucially, position features succeed without any goal representation at all, lifting a raw-goal baseline from 15.7 to 71.0. DGR's learned representation provides an additional $+12.9\;[+9.0, +16.9]$ on top of them, indicating that while goal representations are not useless, their contribution is second-order.
Standardising the coordinates in place, which changes their scale without adding a basis, does not reproduce
the gain and is severely harmful.

\subsection{Ruling Out Confounding Mechanisms}
\label{sec:mechanism-controls}

To understand why state-side intervention drives performance, we systematically evaluated and ruled out several candidate mechanisms:
\begin{itemize}
    \item \textbf{Model capacity:} Widening the actor to $1024$ hidden units changes nothing ($-0.4\;[-8.8, +8.0]$), and the best-performing interface uses fewer parameters ($635\mathrm{k}$) than standard concatenation ($676\mathrm{k}$).
    \item \textbf{Input scale:} Standardising coordinates drops success from $32.2$ to $7.1$, confirming that position requires high-frequency distinguishability rather than standard conditioning.
    \item \textbf{Goal input width:} Compressing $\varphi(g)$ to match the attention context width produces no discernible change ($-2.8\;[-13.5, +7.8]$ with the ideal representation; $-0.1\;[-6.5, +6.3]$ with raw goals).
    \item \textbf{Internal shortest-path routing:} Cross-attention over landmark tokens does not implement min-plus decoding~\eqref{eq:min-plus-decoding}. Attention peaks align with a state-independent goal-proximity heuristic rather than true min-plus argmins.
\end{itemize}

\textbf{Interface corroboration and scope.} Combining state and goal via cross-attention improves antmaze-large performance by $+23.2\;[+15.6, +30.9]$ because queries are formed directly from state observations. Crucially, this gain is redundant when an explicit state code is already present ($+8.9\;[-3.4, +21.2]$). These benefits remain specific to complex navigation: they do not transfer to manipulation tasks or low-dimensional environments like pointmaze, where coordinates already dominate the state vector. We provide complete attention diagnostics, parameter sweeps, and scope evaluations in Appendix~\ref{app:mechanism-controls}.

\section{Conclusion}
\label{sec:conclusion}

We investigated whether goal-representation quality bounds performance in offline goal-conditioned navigation. Across our benchmarks, replacing learned goal embeddings with exact shortest-path representations, degrading their distance structure, or substituting random goal codes does not substantially change downstream task success. In contrast, applying the same interventions to the agent's observation more than doubles success on antmaze-large, and a map-free Fourier positional encoding raises GCIVL from 32 to 84 while roughly doubling performance on humanoidmaze-medium. These findings indicate that the state pathway, rather than the goal representation, is the primary performance bottleneck.

\section{Limitations}
Our findings carry several limitations: while we rule out capacity, input scale, and internal shortest-path routing, the precise mechanism explaining why spatial distinguishability aids policy learning remains an open question. Furthermore, our evaluation is restricted to state-based navigation; the benefits do not transfer to manipulation tasks or low-dimensional environments like pointmaze, where coordinates already dominate observations (Appendix~\ref{app:scope}). Finally, we do not claim state-of-the-art results over hierarchical methods, and position features remain untested in pixel-based domains.

\bibliographystyle{iclr2027_conference}
\bibliography{iclr2027_conference}
\newpage
\appendix

\section{Related Work}
\label{sec:related}

\textbf{Offline goal-conditioned RL.} Goal-conditioned policies trained to reach arbitrary states date back
to \citet{kaelbling1993ijcai-learning} and were scaled with universal value approximators
\citep{3045118.3045258} and hindsight relabelling \citep{3295222.3295258, ghosh2020learningreachgoalsiterated}. The offline setting inherits the distribution-shift problems of offline RL
\citep{levine2020offlinereinforcementlearningtutorial, 3495724.3495824} and is typically addressed with
in-sample value learning \citep{kostrikov2021offlinereinforcementlearningimplicit}. We use the state-based navigation tasks and the two
downstream algorithms of OGBench \citep{ICLR2025_ecd92623}, which also ships oracle goal-representation variants of
these tasks; those variants are reported by \citet{park2026transitiverlvaluelearning} but were not benchmarked when introduced.

\textbf{Goal representations from temporal structure.} The premise we test is that a goal is best described
by its ``reachability'' relations. Contrastive objectives learn such structure from future occupancy
\citep{3600270.3602850, sermanet2018timecontrastivenetworksselfsupervisedlearning}, value-implicit pre-training learns it from
value functions \citep{ma2023vipuniversalvisualreward}, and quasimetric and Hilbert embeddings impose the metric
structure architecturally \citep{wang2023optimalgoalreachingreinforcementlearning, wang2022learninglearnabilityquasimetrics, park2024foundationpolicieshilbertrepresentations, myers2025offlinegoalconditionedreinforcementlearning, NEURIPS2025_dc8fe792}. Dual goal
representations extend this idea, describing a goal by its temporal distances to all other states
\citep{ICLR2026_f8cd7eb1}. The appeal of distance-structured goals is that such distances compose
over the horizon \citep{ICLR2025_bddd4e76}. Our contribution is a
measurement of what this family buys: we construct the exact representation these methods approximate and
find that downstream control is insensitive to it. Other goal or state representation objectives, including
information bottlenecks \citep{alemi2019deepvariationalinformationbottleneck}, self-predictive and contrastive
auxiliary losses \citep{NEURIPS2020_f3ada80d, schwarzer2021dataefficientreinforcementlearningselfpredictive, srinivas2020curlcontrastiveunsupervisedrepresentations}, and behavioural metrics \citep{NEURIPS2021_fd06b8ea},
are the comparison columns we quote from \citet{ICLR2026_f8cd7eb1}.

\textbf{What else has been identified as the bottleneck.} A parallel line of work locates the difficulty in
the horizon rather than the representation, and reduces it with hierarchy or $n$-step returns
\citep{park2024hiqlofflinegoalconditionedrl, 3327144.3327250}, or with explicit search over the replay
buffer \citep{NEURIPS2019_5c48ff18, savinov2018semiparametric, chanesane2021goalconditionedreinforcementlearningimagined}. These methods remain stronger
than ours in absolute terms on the hardest mazes \citep{ICLR2025_ecd92623}. Our finding is complementary: where that literature shortens the
decision problem, we change how the agent's own position enters the network.

\textbf{Input encodings and conditioning.} Random features \citep{2981562.2981710} and Fourier features
\citep{3495724.3496356} are standard tools for making coordinate inputs learnable, and are central to
implicit neural representations \citep{10.1145/3503250, sitzmann2020implicitneuralrepresentationsperiodic}; the usual justification is spectral bias
\citep{rahaman2019spectralbiasneuralnetworks}, which our frequency sweep does not support. How two inputs are combined is itself a
design axis, studied as feature-wise modulation \citep{3504035.3504518}, attention \citep{vaswani2023attentionneed},
and multiplicative interactions \citep{jayakumar2020iclr-multiplicative}, but it is fixed to concatenation
throughout the goal-representation literature. Finally, our emphasis on intervals, seed counts and
pre-registered thresholds follows recommended practice for empirical RL
\citep{henderson2019deepreinforcementlearningmatters, agarwal2022deepreinforcementlearningedge, colas2018randomseedsstatisticalpower}, and our use of a decoder rather
than a trained probe follows the control-task critique of probing \citep{hewitt-liang-2019-designing}.
\section{Experimental Setup and Protocol}
\label{app:setup}

\textbf{Tasks and datasets.} We use the state-based navigation tasks of OGBench \citep{ICLR2025_ecd92623}:
pointmaze-medium, pointmaze-large, antmaze-medium, antmaze-large and humanoidmaze-medium, with the standard
\texttt{navigate} datasets. In these tasks temporal distance is a graph geodesic, which is what makes the
ideal representation computable. Observations are 4, 4, 29, 29 and 69 dimensional, and two dimensions hold the
agent's position in every case.

\textbf{Training and evaluation.} Every run trains for one million gradient steps. We evaluate on the five
held-out goals of the benchmark with fifty episodes each, every 100k steps, and report the mean success rate
over the final three evaluations at 800k, 900k and one million steps. We never report the best evaluation over
training. A single run takes between two and twelve hours on one RTX 4090 or RTX 3090.

\textbf{Reporting.} Arms are compared with Welch's unequal-variance $t$-test and 95\% confidence intervals,
and every table and figure states its seed count. Because several of our central claims are null results, we
report intervals rather than $p$-values alone, following recommended practice for empirical reinforcement
learning \citep{henderson2019deepreinforcementlearningmatters, agarwal2022deepreinforcementlearningedge, colas2018randomseedsstatisticalpower}.

\textbf{Constructing the exact representation.} We sample $K$ landmark states uniformly from the dataset, map
each one to the maze cell that contains it, and run a breadth-first search from every landmark to every cell.
The representation has to be a function of the raw observation rather than a column of the dataset, because at
evaluation time the environment supplies goals that have no dataset index. We therefore store the affine map
from coordinates to cell indices next to the table, so the same lookup serves dataset goals and evaluation
goals. Cells occupied by walls are filled by nearest neighbour, so the lookup is always defined. On
antmaze-large this gives a table over 736 cells drawn from 186 distinct landmark cells, and every state in the
dataset maps to a free cell. The scale of the resulting vector is a free choice that the learned
representation makes implicitly. We make it explicit and standardise each coordinate, and report the
alternatives in Appendix~\ref{app:goal-side}.

\section{Architecture and Implementation Details}
\label{app:architecture}

This section describes the networks, the two downstream algorithms and the inputs we intervene on. Anything
not stated here follows the reference implementation of \citet{ICLR2026_f8cd7eb1}, and all arms within a
comparison share it.

\subsection{Networks}
\label{app:networks}

Every network is a multilayer perceptron with three hidden layers of 512 units, GELU activations and layer
normalisation. The value function is an ensemble of two heads and is trained against a target copy that
tracks the online parameters at rate $\tau = 0.005$. The actor outputs a Gaussian with a state-independent
standard deviation. The learned dual goal representation comes from a separate network of the same shape,
trained on the offline data with a bilinear value parameterisation, and it enters the downstream learner
through a stop-gradient. Goals are supplied by hindsight relabelling with the ratios of the benchmark.
Training uses Adam with a learning rate of $3 \times 10^{-4}$ and a batch size of 1024. Table~\ref{tab:hyperparameters}
lists the values we used.

\begin{table}[h]
\centering
\caption{\textbf{Hyperparameters.} Shared values follow the benchmark. Algorithm-specific and method-specific
values are grouped below them.}
\label{tab:hyperparameters}
\small
\tablebodyfont
\resizebox{\textwidth}{!}{%
\begin{tabular}{ll}
\toprule
Hyperparameter & Value \\
\midrule
Optimiser & Adam \\
Learning rate & $3 \times 10^{-4}$ \\
Batch size & $1024$ \\
Gradient steps & $10^{6}$ \\
Hidden dimensions (value, actor, representation) & $(512, 512, 512)$ \\
Nonlinearity & GELU \\
Layer normalisation & yes \\
Target update rate $\tau$ & $0.005$ \\
Discount $\gamma$ & $0.99$, and $0.995$ on humanoidmaze-medium \\
Representation width $K$ & $256$ on antmaze and humanoidmaze, $64$ on pointmaze \\
Representation parameterisation & bilinear \\
Representation expectile & $0.9$ \\
\midrule
GCIVL value expectile & $0.9$ \\
GCIVL advantage temperature $\alpha$ & $10.0$ \\
GCIVL value goal ratios (current, trajectory, random) & $(0.2, 0.5, 0.3)$ \\
\midrule
CRL critic latent dimension & $512$ \\
CRL behaviour-cloning coefficient $\alpha$ & $0.1$ \\
CRL value goal ratios (current, trajectory, random) & $(0.0, 1.0, 0.0)$ \\
\midrule
Actor goal ratios (current, trajectory, random) & $(0.0, 1.0, 0.0)$ \\
Position features $F$ & $128$, giving 256 appended dimensions \\
Frequency scale $\varsigma$ (cycles per coordinate unit) & $0.025$, $0.1$, $0.4$ \\
\bottomrule
\end{tabular}%
}
\end{table}

\subsection{Downstream Algorithms}
\label{app:algorithms}

\textbf{GCIVL.} GCIVL fits a goal-conditioned value function by expectile regression on the benchmark reward
and extracts a policy by advantage-weighted regression \citep{kostrikov2021offlinereinforcementlearningimplicit}.
With target value $\bar V$ and expectile $\kappa$,
\begin{equation}
\begin{aligned}
\mathcal{L}_V &\;=\; \mathbb{E}_{s, s', g \sim \mathcal{D}}\Bigl[\, \ell_{\kappa}\bigl(r(s,g) + \gamma \bar V(s', \varphi(g)) - V(s, \varphi(g))\bigr) \,\Bigr], \\
\mathcal{L}_{\pi} &\;=\; -\,\mathbb{E}_{s, a, g \sim \mathcal{D}}\Bigl[\, \exp\bigl(\alpha \, A(s, a, g)\bigr) \, \log \pi\bigl(a \mid s, \varphi(g)\bigr) \,\Bigr],
\end{aligned}
\label{eq:gcivl-losses}
\end{equation}
where $\ell_{\kappa}(u) = |\kappa - \mathbf{1}[u < 0]| \, u^{2}$ is the expectile loss, $A$ is the advantage
implied by the value function, and the exponentiated advantage is clipped at 100. We keep the benchmark
implementation, in which the value function is an ensemble of two heads and the expectile weight is taken from
the target advantage.

\textbf{CRL.} CRL fits a critic that is bilinear in a state-action embedding and a goal embedding, and trains
it as a binary classifier that separates goals reached later in the same trajectory from goals taken from
elsewhere in the batch \citep{3600270.3602850}. With $f(s, a, g) = u(s,a)^{\top} v(\varphi(g)) / \sqrt{m}$ over a batch
of size $B$,
\begin{equation}
\begin{aligned}
\mathcal{L}_f &\;=\; -\,\frac{1}{B^{2}} \sum_{i,j} \Bigl[\, \mathbf{1}[i = j] \log \sigma\bigl(f_{ij}\bigr) + \mathbf{1}[i \neq j] \log \bigl(1 - \sigma(f_{ij})\bigr) \,\Bigr], \\
\mathcal{L}_{\pi} &\;=\; -\,\mathbb{E}\Bigl[\, f\bigl(s, \mu(s, \varphi(g)), g\bigr) \,\Bigr] \big/ \mathbb{E}\bigl[|f|\bigr] \;-\; \alpha \, \mathbb{E}\Bigl[\, \log \pi\bigl(a \mid s, \varphi(g)\bigr) \,\Bigr],
\end{aligned}
\label{eq:crl-losses}
\end{equation}
so the policy maximises a scale-normalised critic value with a behaviour-cloning term of weight $\alpha$. The
two algorithms differ in how they estimate value and in how they extract a policy, which lets us separate
effects that are specific to one family from those that are not.

\subsection{The Inputs We Intervene On}
\label{app:inputs}

All five inputs below enter the same networks through the same concatenation interface, unless an experiment
states otherwise. The objective, the optimiser, the network widths and the training length never change.

\textbf{Learned representation.} The dual goal representation of \citet{ICLR2026_f8cd7eb1}, produced by the
bilinear representation network described above. This is the baseline.

\textbf{Ideal representation.} Equation~\eqref{eq:goal-representation} evaluated with exact breadth-first
distances, standardised per coordinate. Nothing about it is learned.

\textbf{Corrupted representations.} The ideal representation with the noise of
equation~\eqref{eq:representation-corruption} added at a fixed seed, either to all entries or to the furthest
quartile only. Each setting is a deterministic representation, not observation noise.

\textbf{State codes.} A fixed vector per maze cell appended to the observation as in
equation~\eqref{eq:state-code}. The code is either the exact distance row, a corrupted row, or a random
$\mathcal{N}(0, 1)$ draw. These require the maze layout and are diagnostics, not a method.

\textbf{Position features.} Equation~\eqref{eq:position-features}. We draw $B \in \mathbb{R}^{F \times 2}$
once from $\mathcal{N}(0, \varsigma^{2})$ with $F = 128$, keep it fixed across seeds, and append the sine and
cosine of the projected coordinates to the observation. This is the standard random Fourier construction
\citep{2981562.2981710, 3495724.3496356} applied to the two observation dimensions that hold the agent's
position. It needs no map and no change to the objective, and the only thing it changes about the agent is the
width of the observation.

\section{Measuring Representation Quality}
\label{app:quality}

When every cell is a landmark, the min-plus decode of equation~\eqref{eq:min-plus-decoding} reproduces the
true geodesic to $9.5 \times 10^{-7}$, which is one unit in the last place of single precision. We use this as
a joint correctness check on the table, the indexing and the decode.

\textbf{Why not a probe.} Regressing temporal distance from a representation is the natural alternative, and
it works in other domains. It does not work here. On antmaze-large a multilayer probe scores between 0.997 and
0.999 on a raw two-dimensional goal, on a frozen random projection and on a learned representation alike, and
a bilinear probe ranks a random 256-dimensional projection above the raw goal. The reason is structural. In a
maze the goal state determines its own position, so any representation that preserves goal identity contains
complete distance information, and the question of how much distance is decodable is saturated by
construction. This is the control-task problem in probing \citep{hewitt-liang-2019-designing}: a probe strong
enough to recover the quantity can recover it from a representation that does not encode it. Min-plus
decoding avoids this because it reads the representation directly and fits nothing.

\section{Additional Goal-Side Results}
\label{app:goal-side}

\subsection{The Full Navigation Suite}
\label{app:ceiling-suite}

\begin{figure}[h]
    \centering
    \includegraphics[width=\textwidth]{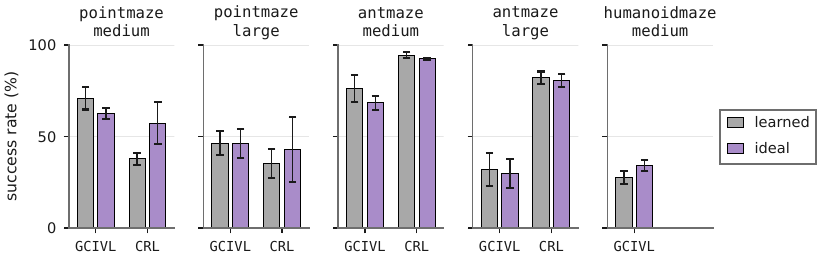}
    \caption{\textbf{Exact and learned goal representations across the full navigation suite.} Mean $\pm$ 1 SD
    over 5 to 8 seeds. CRL was not run on humanoidmaze-medium.}
    \label{fig:ceiling}
\end{figure}

Evaluating downstream performance ceilings across all five environments and both algorithms underscores that representation exactness does not drive control (Figure~\ref{fig:ceiling}; Table~\ref{tab:ideal-vs-learned}).

Across these settings, the exact goal representation ties the learned baseline in most cases. Where differences do appear, they switch signs between algorithms and remain small relative to run-to-run seed variance. Across all five environments, providing exact distance information fails to unlock performance that the learned representation could not already achieve.

\subsection{Corruption Beyond antmaze-large}
\label{app:ladder-other}

Table~\ref{tab:ladder-other} repeats the far-field ladder in three more settings. On antmaze-medium the arms
stay together under both algorithms, and on pointmaze-large under GCIVL the corrupted arms are if anything
slightly above the exact one. None of the three shows a monotone decline with corruption strength.

The exception is pointmaze-large under CRL, where the uncorrupted arm scores $43.0 \pm 17.6$ and the corrupted
arms range from 14.7 to 31.5. This cell has the widest seed variance anywhere in our data, and each corrupted
setting has two seeds. We report it for completeness and do not read a dependence on representation quality
out of it.

\begin{table}[h]
    \centering
    \caption{\textbf{Far-field corruption beyond antmaze-large.} Success rate $\times 100$, mean $\pm$ 1 SD
    over $n$ seeds. Larger $\sigma$ removes more long-range distance structure.}
    \label{tab:ladder-other}
    \small
    \tablebodyfont
    \setlength{\tabcolsep}{4pt}
\resizebox{\textwidth}{!}{%
    \begin{tabular}{llccccc}
        \toprule
        Environment & Algo & Uncorrupted & $\sigma = 1$ & $\sigma = 2$ & $\sigma = 4$ & $\sigma = 8$ \\
        \midrule
        antmaze-medium & GCIVL & $68.4 \pm 4.0$ (5) & $64.9 \pm 3.4$ (2) & $69.2 \pm 2.5$ (2) & $71.5 \pm 5.6$ (2) & $70.7 \pm 0.6$ (2) \\
        antmaze-medium & CRL   & $92.5 \pm 0.4$ (2) & $92.1 \pm 2.0$ (2) & $93.1 \pm 0.4$ (2) & $91.1 \pm 1.6$ (2) & $90.5 \pm 5.2$ (2) \\
        pointmaze-large & GCIVL & $46.3 \pm 7.9$ (5) & $50.7 \pm 4.1$ (2) & $53.4 \pm 7.3$ (2) & $55.0 \pm 4.8$ (2) & $49.3 \pm 0.7$ (2) \\
        pointmaze-large & CRL   & $43.0 \pm 17.6$ (4) & $14.9 \pm 2.7$ (2) & $14.7 \pm 7.2$ (2) & $29.5 \pm 12.1$ (2) & $31.5 \pm 4.9$ (2) \\
        \bottomrule
    \end{tabular}%
}
\end{table}

\subsection{What the Two Corruptions Do to the Geometry}
\label{app:corruption-geometry}

\begin{figure}[h]
    \centering
    \includegraphics[width=0.86\textwidth]{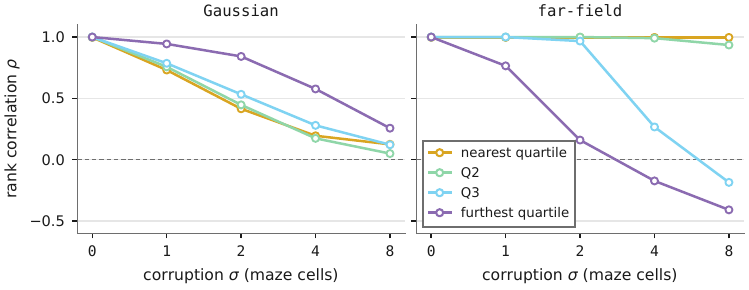}
    \caption{\textbf{Effect of the two corruption procedures on temporal-distance structure.} Spearman
    correlation between the decoded and true distance, within quartiles of true distance. Gaussian corruption
    degrades all ranges; far-field corruption keeps the nearest quartile exact and reverses the furthest.}
    \label{fig:corruption-geometry}
\end{figure}

The two corruption families have differing effects on the goal and state representation. This is portrayed in
Figure~\ref{fig:corruption-geometry}. We decode the distance with
equation~\eqref{eq:min-plus-decoding} and correlate it against the true geodesic within each quartile of true
distance. Gaussian corruption degrades every range at once. Far-field corruption leaves the nearest quartile
exact at every level while driving the furthest quartile negative, so distant goals end up ranked in reverse.
The two therefore damage the representation in different ways, and they arrive at the same place downstream
(Figure~\ref{fig:ladder}). 
% This is what makes the null in Section~\ref{sec:saturated} hard to explain as a representation that happens to survive one particular kind of noise.

\subsection{Normalisation of the Exact Representation}
\label{app:normalisation}

The scale of the exact distance vector is a free choice.
Table~\ref{tab:normalisation} compares the scheme used everywhere else in the paper against unit length and
against raw step counts. Length normalisation is worth more to both algorithms than the difference between the
exact and the learned representation. This is the same theme as the state-side result: how the vector is
presented to the network matters more than the geometry inside it.

\begin{table}[h]
\centering
\caption{\textbf{Normalisation of the exact table on antmaze-large.} Differences are against per-coordinate
standardisation, with a 95\% Welch interval.}
\label{tab:normalisation}
\small
\tablebodyfont
\setlength{\tabcolsep}{3pt}
\begin{tabular}{lccccc}
\toprule
Algo & Standardised & Unit $\ell_2$ & Difference & Raw counts & Difference \\
\midrule
GCIVL & $30.0 \pm 8.0$ (8) & $41.8 \pm 8.1$ (3) & $+11.8\;[-4.2, +27.8]$ & $28.7 \pm 8.5$ (3) & $-1.3\;[-18.2, +15.5]$ \\
CRL & $80.8 \pm 3.4$ (8) & $86.2 \pm 2.9$ (3) & $+5.5\;[-0.1, +11.0]$ & $74.9 \pm 6.1$ (3) & $-5.9\;[-19.3, +7.6]$ \\
\bottomrule
\end{tabular}
\end{table}

\subsection{Number of Landmarks}
\label{app:landmarks}

To test whether the width of the representation matters, we halve and quadruple the landmark count. This leaves
control unchanged in every case we ran (Table~\ref{tab:landmarks}). This also rules out landmark count as the
reason the state-side and interface effects appear on antmaze but not on pointmaze, since pointmaze at $K = 256$
behaves like pointmaze at $K = 64$.

\begin{table}[h]
\centering
\caption{\textbf{Landmark count with the ideal representation.} Concatenation interface. Differences are the
larger count minus the smaller, with a 95\% Welch interval.}
\label{tab:landmarks}
\small
\tablebodyfont
\setlength{\tabcolsep}{4pt}
\begin{tabular}{llccc}
\toprule
Environment & Algo & $K = 64$ & $K = 256$ & Difference \\
\midrule
antmaze-large & GCIVL & $28.5 \pm 6.9$ (5) & $30.0 \pm 8.0$ (8) & $+1.5\;[-7.9, +10.9]$ \\
antmaze-large & CRL & $82.6 \pm 4.1$ (3) & $80.8 \pm 3.4$ (8) & $-1.8\;[-10.1, +6.4]$ \\
pointmaze-medium & GCIVL & $62.7 \pm 3.0$ (5) & $66.7 \pm 6.0$ (6) & $+4.0\;[-2.6, +10.6]$ \\
pointmaze-large & GCIVL & $46.3 \pm 7.9$ (5) & $49.0 \pm 6.7$ (7) & $+2.6\;[-7.4, +12.7]$ \\
\bottomrule
\end{tabular}
\end{table}

\section{Additional State-Side Results}
\label{app:state-side}

\subsection{Results the Main Text Refers To}
\label{app:state-extras}
Several additional experiments clarify how and where state-side interventions help, as summarised in Table~\ref{tab:state-extras}.

Three results stand out. First, state codes generalise beyond GCIVL: applying them to CRL improves performance by $+9.2$ points on the same task where goal-side intervention achieved nothing. Second, reading the ideal goal representation with cross-attention across both the actor and value networks adds $+33.2$ points, but only $+3.9$ with the learned representation. Downstream control can exploit higher-quality goal information only when given an interface capable of extracting it. Third, adding cross-attention to an agent that already has an explicit state code yields only $+8.9$ points (not distinguishable from zero), compared to $+23.2$ points when no state code is present. The two interventions provide redundant benefits.
\begin{table}[h]
\centering
\caption{\textbf{State-side and interface results referenced in the main text.} All on antmaze-large.
Differences are against the baseline column with a 95\% Welch interval; bold marks intervals excluding zero.}
\label{tab:state-extras}
\small
\tablebodyfont
\setlength{\tabcolsep}{4pt}
\resizebox{\textwidth}{!}{%
\begin{tabular}{lccc}
\toprule
Configuration & Baseline & With intervention & Difference \\
\midrule
CRL, exact state code & $80.8 \pm 3.4$ (8) & $89.9 \pm 1.5$ (5) & $\boldsymbol{+9.2\;[+6.1, +12.3]}$ \\
GCIVL, state code $+$ cross-attention & $64.7 \pm 9.4$ (5) & $73.7 \pm 7.0$ (5) & $+8.9\;[-3.4, +21.2]$ \\
GCIVL, ideal $+$ x-attn on value and actor & $30.0 \pm 8.0$ (8) & $63.2 \pm 5.0$ (5) & $\boldsymbol{+33.2\;[+25.2, +41.1]}$ \\
GCIVL, learned $+$ x-attn on value and actor & $32.2 \pm 8.9$ (8) & $36.1 \pm 3.3$ (5) & $+3.9\;[-3.9, +11.7]$ \\
\bottomrule
\end{tabular}%
}
\end{table}

\subsection{Which Wavelengths Work}
\label{app:frequency}

\begin{figure}[h]
\centering
\includegraphics[width=0.92\textwidth]{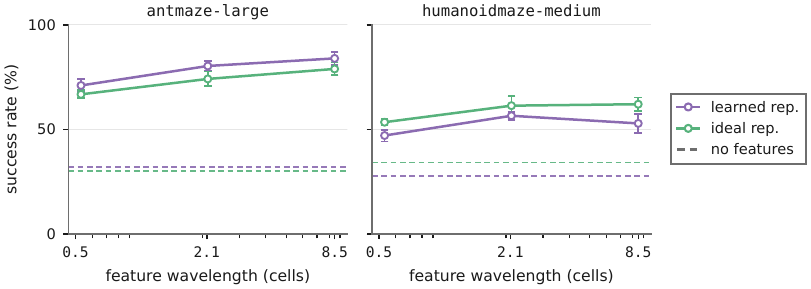}
\caption{\textbf{Success against feature wavelength.} The runs of
Figure~\ref{fig:position-features}, plotted against wavelength in maze cells. Dashed lines in the matching
colour mark the same agent without position features.}
\label{fig:frequency-trend}
\end{figure}

Performance varies systematically with feature wavelength in all four settings (Figure~\ref{fig:frequency-trend}).

Longer wavelengths consistently perform better, peaking at roughly eight maze cells. Short wavelengths act like random codes; they separate states but drop all spatial proximity. In contrast, long wavelengths keep nearby states similar while separating distant ones, which helps the policy most. Because our sweep stopped at eight cells, where this trend eventually peaks remains open.

\section{Interface Definitions}
\label{app:interfaces}

Concatenation is the baseline used by all prior work, equation~\eqref{eq:concat-interface}. FiLM computes
modulation parameters from the goal and applies them to a state-only trunk \citep{3504035.3504518},
\begin{equation}
h \;\leftarrow\; h \odot \bigl(1 + \gamma(\varphi)\bigr) + \beta(\varphi).
\label{eq:film-interface}
\end{equation}
Cross-attention treats the $K$ coordinates of $\varphi$ as tokens. Token $i$ is a learned landmark embedding
added to an embedding of the scalar $\varphi_i(g)$, a query is formed from the state, and the attended context
is concatenated with the state before the trunk \citep{vaswani2023attentionneed},
\begin{equation}
\begin{aligned}
c &\;=\; \mathrm{Attn}\bigl(q(s), \; \{\, e_i + W \varphi_i(g) \,\}_{i=1}^{K}\bigr), \\
I_{\mathrm{xattn}}(s, \varphi) &\;=\; \mathrm{MLP}\bigl([\, s \, ; \, c \,]\bigr).
\end{aligned}
\label{eq:xattn-interface}
\end{equation}
We use four heads and a width of 64 for the tokens and the context. The bottleneck control compresses
$\varphi$ to the same width with a plain network and then concatenates, so any gain that cross-attention gets
merely by narrowing the goal input should appear there as well.

Two properties keep the comparison fair. First, both new interfaces start as a state-only function: the FiLM
modulation begins at the identity because $\gamma$ and $\beta$ are zero-initialised, and the attention output
projection is zero-initialised. Any gain they obtain therefore comes from learning to use the goal and not
from a different function at initialisation. Concatenation and the bottleneck pass the goal through a randomly
initialised first layer, as in prior work. Second, the interfaces are close in size, at 676k parameters for
concatenation, 635k for cross-attention, 660k for the bottleneck and 1.47M for FiLM, so the best interface is
not the largest.

\section{Full Interface Sweep}
\label{app:interface-sweep}

\begin{figure}[h]
\centering
\includegraphics[width=\textwidth]{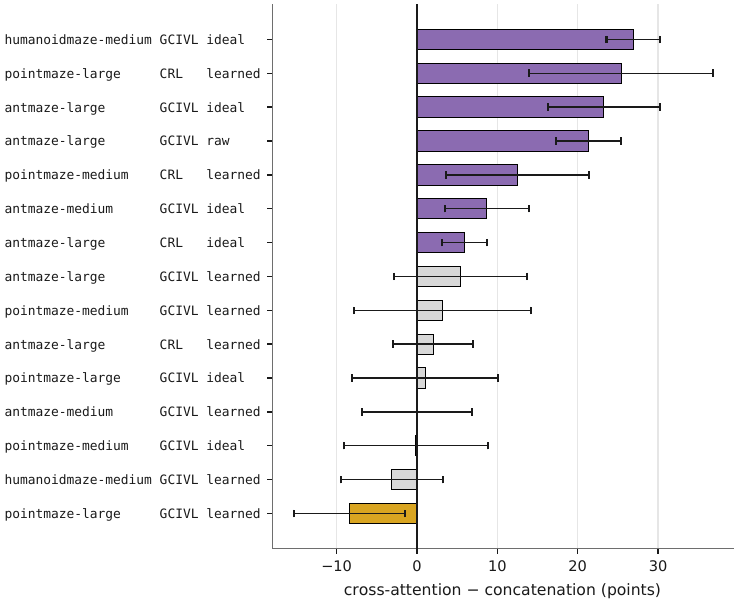}
\caption{\textbf{Full interface sweep.} Cross-attention minus concatenation, with everything else held fixed
within each row. Whiskers are 95\% Welch intervals. Violet marks intervals excluding zero in favour of
cross-attention, gold in favour of concatenation, grey a tie.}
\label{fig:interface-sweep}
\end{figure}

How the policy ingests goal representations often matters more than the representation itself (Figure~\ref{fig:interface-sweep}). Across fifteen controlled comparisons holding representations, algorithms, and hyperparameters fixed, replacing standard concatenation with cross-attention shifts performance from $-8.4$ to $+26.9$ points, with six runs cleanly separated from zero (five favoring cross-attention). Rather than advocating cross-attention as a standalone technique, this comparison demonstrates that standard architectural defaults, often fixed by convention, introduce larger performance shifts than optimising goal representations.

\section{Mechanistic Controls}
\label{app:mechanism-controls}

This section rules out the explanations of the state-side gain that do not involve spatial distinguishability.

\subsection{Capacity, Scale and Input Width}
\label{app:capacity-scale}

\textbf{Capacity.} Doubling the actor width from 512 to 1024 units per hidden layer under default GCIVL
changes nothing, at $-0.4\;[-8.8, +8.0]$. Across the interfaces, parameter count does not track performance
either: the best variant has 635k parameters against 676k for concatenation, and the largest at 1.47M is the
worst.

\textbf{Coordinate scale.} If the gain came from numerical conditioning, then fixing the scale of the
coordinates should be enough. It is not. Standardising $(x, y)$ to zero mean and unit variance, which is the
smallest intervention that aligns position scale with the proprioceptive dimensions, drops success from 32.2 to
7.1. The features help by separating positions, not by rescaling them.

\textbf{Input width.} A high-dimensional goal vector might help simply by being wide. Passing $\varphi(g)$
through a linear bottleneck matched to the attention context width changes nothing, both with the ideal
representation at $-2.8\;[-13.5, +7.8]$ and with raw goal observations at $-0.1\;[-6.5, +6.3]$.

\subsection{The Policy Does Not Route Through Landmarks}
\label{app:attention-analysis}

Cross-attention over landmark tokens is structurally able to carry out the min-plus decode of
equation~\eqref{eq:min-plus-decoding}, so we checked whether it does. We extracted the attention distribution
over 256 landmark tokens for 4096 evaluation state-goal pairs from two independently trained seeds.

Attention puts $1.59\times$ and $1.43\times$ chance mass on the landmark that the true min-plus decode
selects, which looks supportive until compared with a null. A state-independent baseline that always picks the
landmark nearest the goal receives more mass, at $2.52\times$ and $1.85\times$ chance. The min-plus path
length at the attention peak is 9.0 and 11.8 cells, against 5.4 cells at the true argmin and 11.3 cells for a
random landmark. The policy is attending to the goal region, not computing shortest paths through it.

\subsection{Cross-Attention and the State Code Overlap}
\label{app:interface-and-scope}

Cross-attention builds its query from the agent's state, so it introduces a state-dependent positional
encoding implicitly. That predicts an overlap with the explicit state code, and we see one. Cross-attention is
worth $+23.2\;[+15.6, +30.9]$ on antmaze-large on its own, and $+33.2\;[+25.2, +41.1]$ when extended to the
critic, but only $+8.9\;[-3.4, +21.2]$ once a state code is already present (Table~\ref{tab:state-extras}).
The two are largely different routes to the same quantity.

\section{Where the Interventions Do Not Help}
\label{app:scope}
\begin{figure}[h]
\centering
\includegraphics[width=0.7\textwidth]{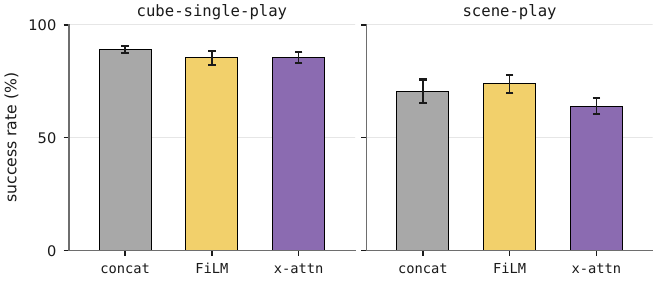}
\caption{\textbf{Interfaces on manipulation tasks.} GCIVL with the learned representation. Cross-attention is
worse than concatenation on cube-single-play, $-3.6\;[-6.5, -0.7]$, and no better on scene-play,
$-6.5\;[-13.2, +0.2]$.}
\label{fig:manipulation}
\end{figure}
\textbf{Pointmaze.} None of the interventions help. Cross-attention with the ideal representation is worth
$-0.2\;[-12.1, +11.8]$ on pointmaze-medium and $+1.0\;[-9.7, +11.8]$ on pointmaze-large. Raising the landmark
count from 64 to 256, the width used on antmaze, does not change that: the gains become
$+1.8\;[-7.2, +10.8]$ and $-2.2\;[-9.9, +5.5]$. Pointmaze observations are four dimensional and the agent's
coordinates are most of them, so there is little for a position encoding to disambiguate.

\textbf{Manipulation.} The interface result does not transfer either. Figure~\ref{fig:manipulation} reports
GCIVL with the learned representation on the two manipulation tasks we ran. Cross-attention is significantly
worse than concatenation on cube-single-play and no better on scene-play. These tasks have no maze geometry
and no single pair of coordinates that locates the agent, which is the structure the interventions exploit.

% \textbf{What we do not claim.} These are controlled comparisons that isolate where the information has to
% enter. They are not a claim of state-of-the-art performance against specialised hierarchical methods
% \citep[e.g.,][]{park2024hiqlofflinegoalconditionedrl}.

\end{document}